\documentclass[runningheads]{llncs}

\usepackage[mobile]{eccv}

\usepackage{eccvabbrv}

\usepackage{graphicx}
\usepackage{booktabs}
\usepackage[font=scriptsize]{caption}

\usepackage[accsupp]{axessibility}  

\newcommand{\xhdr}[1]{\vspace{0em}\noindent{{\bf #1.}}}

\newif\ifcomments
\ifcomments
    \providecommand{\abhinav}[2][]{{\protect\color{magenta}{[\textbf{A}:\textbf{#1} #2]}}}
    \providecommand{\tarun}[2][]{{\protect\color{red}{[\textbf{T}:\textbf{#1} #2]}}}
    \providecommand{\silky}[2][]{{\protect\color{violet}{[\textbf{S}:\textbf{#1} #2]}}}
\else
    \providecommand{\abhinav}[2][]{}
     \providecommand{\tarun}[2][]{}
     \providecommand{\silky}[2][]{}
\fi

\usepackage{hyperref}

\usepackage{orcidlink}

\begin{document}

\title{Towards Efficient Exemplar Based Image Editing with Multimodal VLMs} 



\authorrunning{A. Jadhav, A. Srivastava et al.}


\author{Avadhoot Jadhav\inst{1\textsuperscript{*}} \and
Ashutosh Srivastava\inst{2\textsuperscript{*}} \and
Abhinav Java\inst{3\textsuperscript{†}} \and
Silky Singh\inst{4\textsuperscript{†}} \and
Tarun Ram Menta\inst{5} \and
Surgan Jandial\inst{6\textsuperscript{†}} \and
Balaji Krishnamurthy\inst{5}}

\institute{
Indian Institute of Technology, Bombay \and
Indian Institute of Technology, Roorkee \and
Microsoft Research \and
Stanford University \and
Adobe MDSR \and
Carnegie Mellon University
}

\renewcommand{\thefootnote}{\fnsymbol{footnote}}
\footnotetext{\textsuperscript{*} Work done during internship at Adobe MDSR}
\footnotetext{\textsuperscript{†} Work done while at Adobe}

\newcommand{\method}{\textsc{ReEdit}\xspace}

\maketitle
\vspace{-0.2cm}

\vspace{-0.3cm}
\begin{abstract}
     Text-to-Image Diffusion models have enabled a wide array of image editing applications. However, capturing all types of edits through \textit{text} alone can be challenging and cumbersome. The ambiguous nature of certain image edits is better expressed through an exemplar pair, i.e., a pair of images depicting an image \textit{before} and \textit{after} an edit respectively. In this work, we tackle \textit{exemplar-based image editing} -- the task of transferring an edit from an exemplar pair to a content image(s), by leveraging pretrained text-to-image diffusion models and multimodal VLMs. Even though our end-to-end pipeline is optimization-free, our experiments demonstrate that it still outperforms baselines on multiple types of edits while being $\approx$4x faster.
     \vspace{-0.2cm}
\end{abstract}
\vspace{-0.2cm}


\vspace{-0.65cm}
\section{Introduction}
\label{sec:intro}



\vspace{-0.1cm}
The field of \emph{image editing}~\cite{karras2019style, alaluf2022third, liu2022self, gal2022stylegan, nichol2021glide} focuses on performing desired manipulations to an image. More recently, diffusion models have enabled controllable and precise image editing with text descriptions~\cite{zhang2023adding, mokady2023null, gal2022image, zhang2023sine, huang2023reversion,  kim2022diffusionclip, parmar2023zero, tumanyan2023plug, yang2023object, kwon2022diffusion, couairon2022diffedit, meng2021sdedit, kawar2023imagic, hertz2022prompt, parmar2023zero, brooks2023instructpix2pix, tumanyan2023plug}. The focus of this work is \emph{exemplar-based image editing}. Given a pair of images (or, exemplar pair), we aim to learn an edit such that it can subsequently be applied to an entire corpus of images. This formulation is motivated by the "visual prompting" proposed in~\cite{bar2022visual}. Existing works in this area~\cite{jeong2024visual, yang2023paint, nguyen2024visual, hertzmann2023image, vsubrtova2023diffusion, liao2017visual, yang2024imagebrush} typically optimize a text embedding which is time intensive~\cite{nguyen2024visual}, utilize sophisticated diffusion models like Instruct-pix2pix~\cite{brooks2023instructpix2pix} trained on the specific task of image editing~\cite{nguyen2024visual, hertzmann2023image, vsubrtova2023diffusion}, and can only capture a limited type of edits (for \eg, VISII~\cite{nguyen2024visual} performs extremely well for \emph{global style transfer} type edits). On the contrary, we devise an end-to-end optimization-free method that leverages a vanilla text-to-image diffusion model (Stable Diffusion~\cite{rombach2022high}) and multimodal VLMs (LLaVA~\cite{liu2024llava}) to edit a content image given only an exemplar pair. As a result, our approach is both faster and more effective compared to the baselines, and independent of the base diffusion model. To summarize, the contributions of our work are listed below:




\begin{enumerate}
    \item We propose an inference-time approach for \emph{exemplar-based image editing} that does not require finetuning or optimizing any part of the pipeline. Compared to the most optimal baseline, the runtime of our method is $\approx$4x faster. 

    \item We collate a dataset of 170 exemplar pairs ($x, x_{\text{edit}}$), corresponding test images, covering a wide range of edits. Due to a lack of standard exemplar-based datasets, this is a tangible contribution towards standardized evaluation of \emph{exemplar-based image editing} approaches.

    \item Our rigorous qualitative and quantitative analysis shows that our method performs well on variety of edits, while preserving the structure of the original image. These observations are corroborated by significant improvements in quantitative scores over baselines.
\end{enumerate}
\vspace{-0.2cm}

\vspace{-0.45cm}
\section{Our Methodology: \method}
\label{sec:method}

\begin{figure}[!th]
    \centering
    \includegraphics[width=0.8\textwidth]{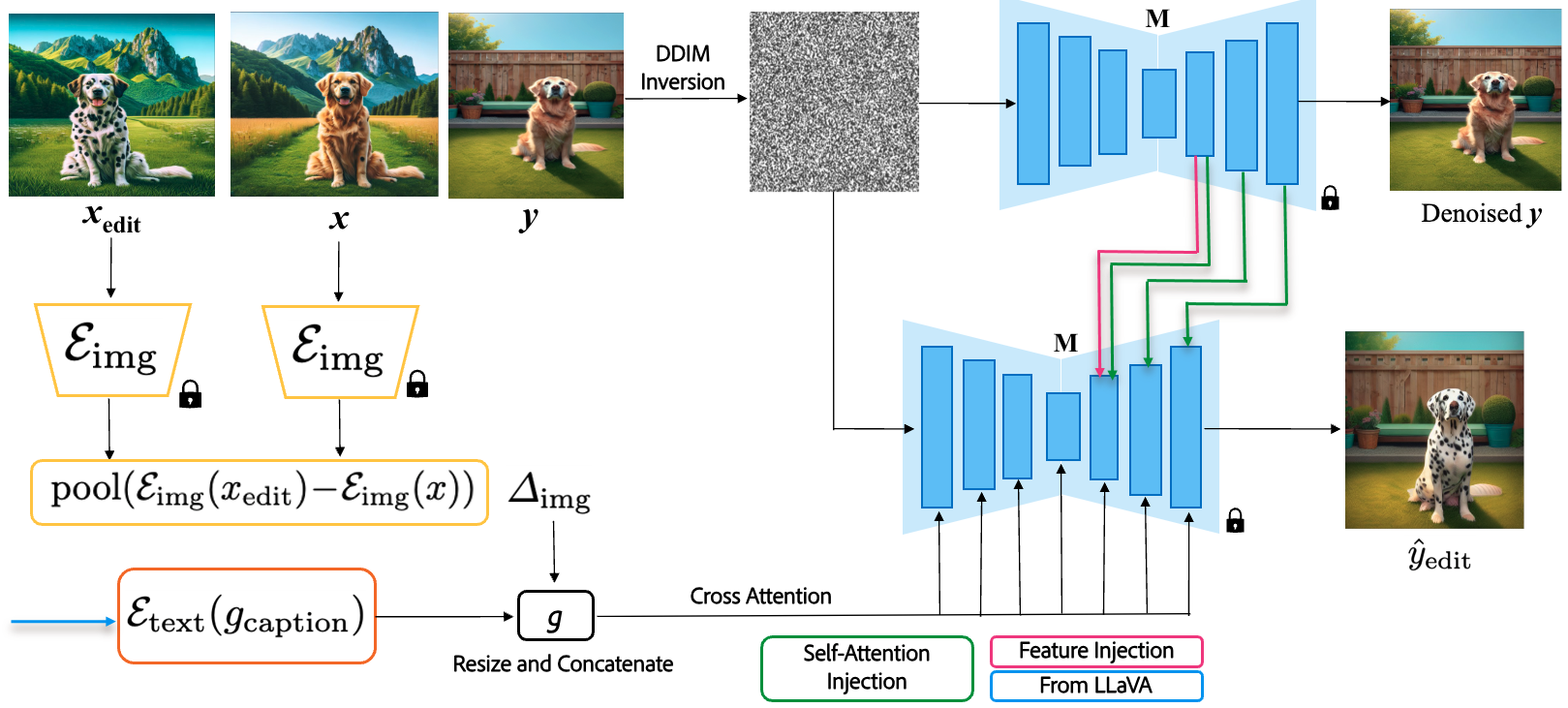}
    \small{\caption{Overview of our framework \method. For details, please refer Section~\ref{sec:method}.\label{fig:overview}}}
\end{figure}

\vspace{-0.2cm}
\noindent{\xhdr{Problem Setting}} Given a pair of exemplar images $(x, x_{\text{edit}})$, where $x$ denotes the original image, and $x_{\text{edit}}$ denotes the edited image respectively. Our objective is to capture the edit (say $g$, such that $x_{\text{edit}} = g(x)$), and apply the \emph{same edit} ($g$) on a test image $y$ to obtain the corresponding edited image $\hat{y}_{\text{edit}}$. We use $M$ to denote pretrained diffusion model (Stable Diffusion~\cite{rombach2022high}). Our approach comprises two key steps as explained below with an overview in Fig.~\ref{fig:overview}.


\vspace{-0.3cm}
\vspace{-0.3cm}


\vspace{-0.3cm}
\noindent{\xhdr{a. Capturing Edits from exemplars}}
We posit that \emph{textual descriptions are necessary but not sufficient} to generate $\hat{y}_{\text{edit}}$ from ($x, x_{\text{edit}}, y$). Consequently, we capture edits in both \textit{text} and \textit{image} space. Firstly, we leverage a multimodal VLM (LLaVA~\cite{liu2024improved, liu2024llava, liu2024visual}) to verbalize the edits in the exemplar pair ($x$, $x_{\text{edit}}$). We pass these images as a grid, along with a detailed prompt $p_1$ that instructs LLaVA to generate a comprehensive description of the edits, denoted by $g_{\text{text}}$. Additionally, to provide the context of the test image $y$, we curate another prompt $p_2$ instructing LLaVA to describe $\hat{y}_{\text{edit}}$ in text after applying the edit $g_{\text{text}}$ on $y$. As a result, we obtain a final text description of $\hat{y}_{\text{edit}}$, denoted by $g_{\text{caption}}$. To reduce verbosity and token length, we limit $g_{\text{caption}}$ to $20$ words.

Note that text alone still can't capture the specific style, intensity, hue, saturation, exact shape, or other detailed attributes of the objects in the image. Therefore, we also capture the edits from $(x, x_{\text{edit}})$ directly in CLIP's embedding space: $\Delta_{\text{img}} = \operatorname{pool}(\mathcal{E}_{\text{img}}(x_{\text{edit}}) - \mathcal{E}_{\text{img}}(x))$, where $\mathcal{E}_{\text{img}}$ is the CLIP image encoder, and $\operatorname{pool}$ denotes an operator that resizes the output to the desired size for $M$. Our final edit embedding is thus given by $g := \operatorname{concat}(\Delta_{\text{img}},~ \mathcal{E}_{\text{text}}(g_{\text{caption}}))$, $\mathcal{E}_{\text{text}}$ is CLIP text encoder. Both the image and text conditioning in $g$ work in tandem to provide nuanced guidance for precise edits.

\vspace{0.15cm}
\noindent{\xhdr{b. Conditioning Stable Diffusion on $(g, y)$}}
A crucial requirement of image editing approaches is that they preserve the content and structure of the original image in the edited output. Thus, we aim to condition $M$ on $g$ such that only the relevant parts of $y$ are edited, while the rest of the image remains intact. To achieve this, we introduce the approach of attention and feature injection motivated by~\cite{tumanyan2023plug}. Specifically, we invert $y$ using DDIM inversion~\cite{song2020denoising}, and run vanilla denoising on the inverted noise ($y_{\text{noise}}$) to collect its features (say $f$) and attention matrices ($Q, K, V$). These features from upsampling blocks contain the overall structure information for $y$~\cite{tumanyan2023plug}. In a parallel run, starting with $y_{\text{noise}}$, we condition the denoising process on edit $g$ (through cross-attention), inject the features ($f$) at the fourth layer and modify the keys and queries ($Q, K$) in the self-attention layers (from layers $4$ to $11$) of $M$ to obtain $\hat{y}_{\text{edit}}$.

\vspace{-0.2cm}
\section{Experiments and Results}
\label{sec:experiments}
\vspace{-0.3cm}

\noindent{\xhdr{Setup and Evaluation}} Our method is an inference-time approach, and is directly applicable to an arbitrary set of $(x, x_{\text{edit}}, y)$ images. However, there are no standard exemplar-based datasets for evaluation. To this end, we curate a dataset of $170$ examples ($x, x_{\text{edit}}, y$) and their corresponding ground truth image $y_{\text{edit}}$ using several existing datasets from Instruct-Pix2Pix~\cite{brooks2023instructpix2pix}, HQ-Edit~\cite{hui2024hq} and Imagic~\cite{kawar2023imagic}. We prepare a strong set of baselines by making important additions to VISII~\cite{nguyen2024visual} (VISII, VISII w/ LLaVA edit text). Additionally, we also benchmark against text-based editor Instruct-pix2pix~\cite{brooks2023instructpix2pix} (IP2P), by generating a short edit instruction using LLaVA. Finally, we compute a total of $7$ metrics measuring structural and perceptual similarity between $(\hat{y}_{\text{edit}}, y_{\text{edit}})$ (LPIPS~\cite{zhang2018unreasonable}, SSIM~\cite{wang2004image}, FID~\cite{heusel2017gans}), faithfulness of the edit with the exemplar pair (CLIP score~\cite{hessel2021clipscore}, Dir. Similarity~\cite{gal2021stylegan}, S-Visual~\cite{nguyen2024visual}), and a Human Preference Score (HPS)~\cite{wu2023human}. 

\vspace{-0.3cm}
\begin{table}[!hb]
\centering
\scalebox{0.7}{
\begin{tabular}{@{}lcccc@{}}
\toprule
\textbf{Metric} & \begin{tabular}[c]{@{}c@{}}IP2P w/ LLaVA \\ edit text~\cite{brooks2023instructpix2pix}\end{tabular} & \begin{tabular}[c]{@{}c@{}}VISII w/ LLaVA\\ edit text\end{tabular} & \begin{tabular}[c]{@{}c@{}}VISII~\cite{nguyen2024visual}\end{tabular} & \textbf{\method} (Ours) \\ \midrule

LPIPS~\cite{zhang2018unreasonable} ($\downarrow$) & $0.28\pm0.6714$ & \underline{$0.20\pm0.8281$} & $0.37\pm0.5072$ & \textbf{0.16}$\pm$\textbf{0.4812} \\

FID~\cite{heusel2017gans} ($\downarrow$) & $1.3097$ & \textbf{0.3049} & $1.2365$ & \underline{$0.5346$} \\

HPS~\cite{wu2023human} ($\uparrow$) & \textbf{0.22}$\pm$\textbf{0.1493} & \textbf{0.22}$\pm$\textbf{0.1446} & \underline{$0.21\pm0.1614$} & \textbf{0.22}$\pm$\textbf{0.1460} \\

SSIM~\cite{wang2004image} ($\uparrow$) & $0.56\pm0.3128$ & \underline{0.62$\pm$0.2694} & $0.48\pm0.3376$ & \textbf{0.63}$\pm$\textbf{0.1925} \\

CLIP Score~\cite{hessel2021clipscore} ($\uparrow$) & $23.78\pm0.1509$ & $23.93\pm0.1465$ & \textbf{24.74}$\pm$\textbf{0.1347} & \underline{$24.56\pm0.1273$} \\

\begin{tabular}[c]{@{}l@{}}Dir. Similarity~\cite{gal2021stylegan} ($\uparrow$) \end{tabular} & \underline{0.03$\pm$1.4923} & \textbf{0.04}$\pm$\textbf{1.3783} & \textbf{0.04}$\pm$\textbf{1.3188} & \textbf{0.04}$\pm$\textbf{1.1361} \\

S-Visual~\cite{nguyen2024visual} ($\uparrow$) & $0.09\pm1.846$ & \underline{$0.10\pm1.5652$} & \textbf{0.16}$\pm$\textbf{0.8589} & $0.09\pm1.7038$ \\ \bottomrule
\end{tabular}}
\vspace{0.05cm}
\caption{Quantitative comparison of our framework \method against strong baselines -- VISII (and its modifications), and Instruct-pix2pix (IP2P). Reported are the mean and coefficient of variance of $7$ different metrics on our dataset. (best scores in \textbf{bold}; second best \underline{underlined})} \label{tab:quant}
\end{table}





\vspace{-0.5cm}
\noindent{\xhdr{Quantitative \& Qualitative analysis}} 
As shown in Table~\ref{tab:quant}, our framework \method performs favorably to baselines on HPS, SSIM, CLIP Score, and Dir. Similarity metrics; while significantly improves on LPIPS score (achieving $0.16$ compared to second best score of $0.20$ -- VISII w/ LLaVA edit text). We atttribute the relatively high score of VISII~\cite{nguyen2024visual} on S-Visual metric to the presence of this term in their optimization process itself. The overall runtime of our approach \textbf{($\approx$ 140 sec)} is much faster than VISII which takes upwards of 9 minutes. 

\begin{figure}[!th]
    \centering
    \begin{minipage}[t]{0.13\textwidth}
        \centering
        \subcaption[]{$x$}{}
        \includegraphics[width=\textwidth]{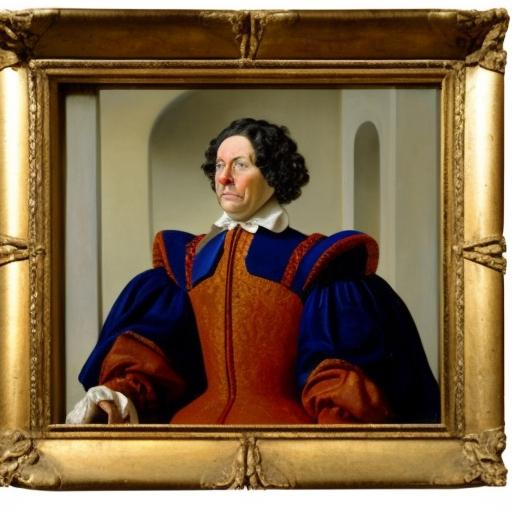}
    \end{minipage}%
    \begin{minipage}[t]{0.13\textwidth}
        \centering
        \subcaption[]{$x_{\text{edit}}$}{}
        \includegraphics[width=\textwidth]{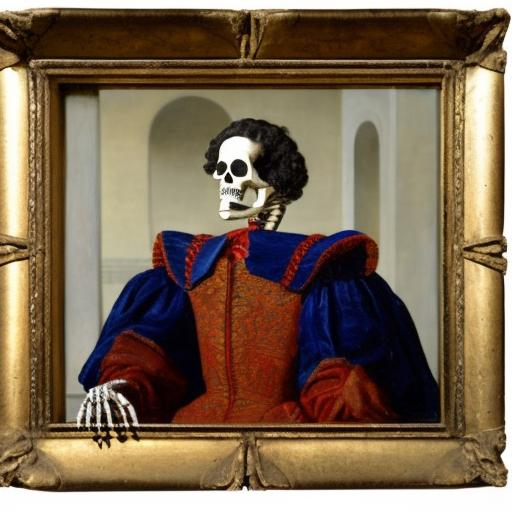}
    \end{minipage}%
    \begin{minipage}[t]{0.13\textwidth}
        \centering
        \subcaption[]{$y$}
        \includegraphics[width=\textwidth]{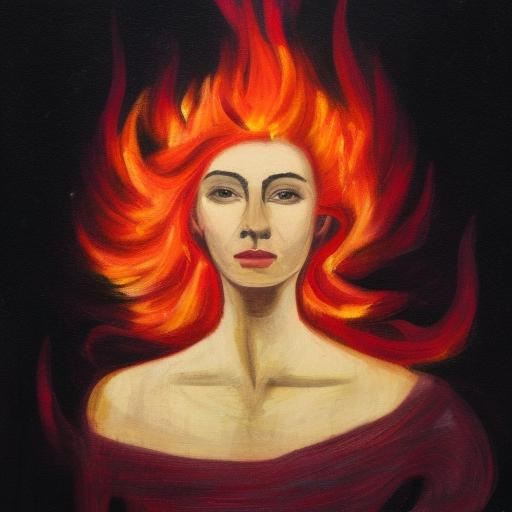}
    \end{minipage}%
    \begin{minipage}[t]{0.13\textwidth}
        \centering
        \subcaption[]{\textbf{\method}}{}
        \includegraphics[width=\textwidth]{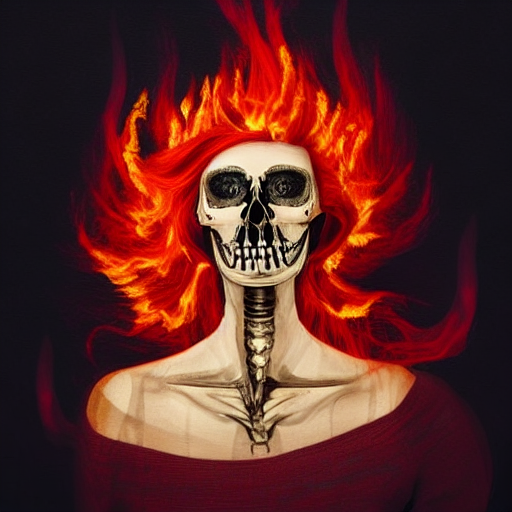}
    \end{minipage}%
    \begin{minipage}[t]{0.13\textwidth}
        \centering
        \subcaption[]{VISII}{}
        \includegraphics[width=\textwidth]{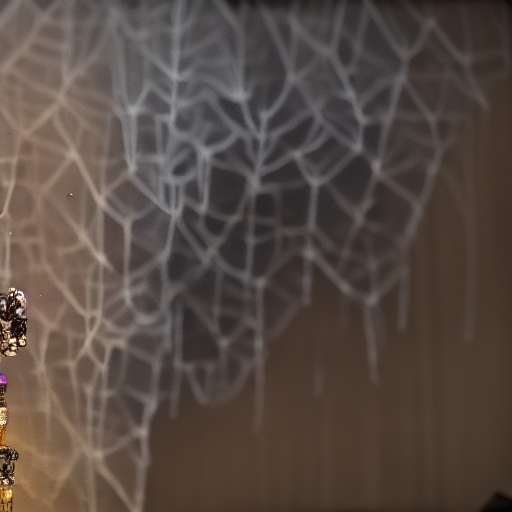}
    \end{minipage}%
    \begin{minipage}[t]{0.13\textwidth}
        \centering
        \subcaption[]{VISII Text}{}
        \includegraphics[width=\textwidth]{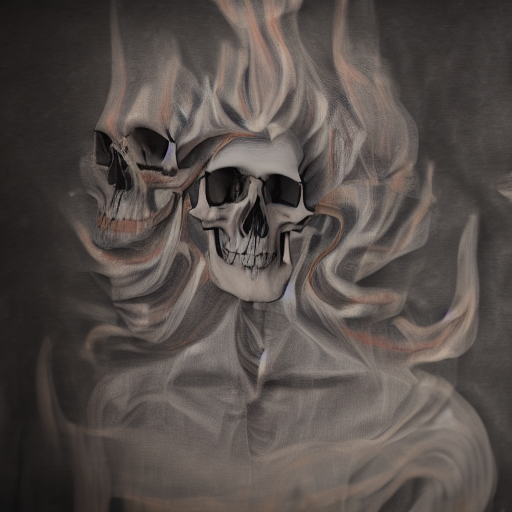}
    \end{minipage}%
    \begin{minipage}[t]{0.13\textwidth}
        \centering
        \subcaption[]{IP2P Text}{}
        \includegraphics[width=\textwidth]{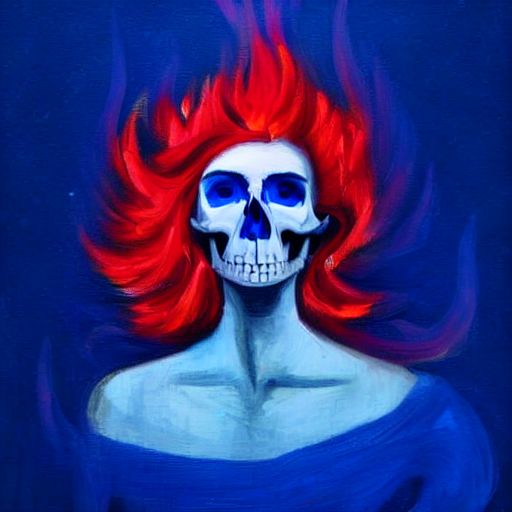}
    \end{minipage}%
    \par

    \begin{minipage}[t]{0.13\textwidth}
        \centering
        \includegraphics[width=\textwidth]{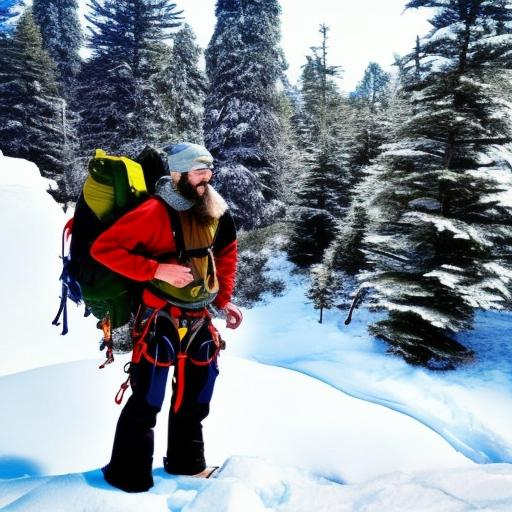}
    \end{minipage}%
    \begin{minipage}[t]{0.13\textwidth}
        \centering
        \includegraphics[width=\textwidth]{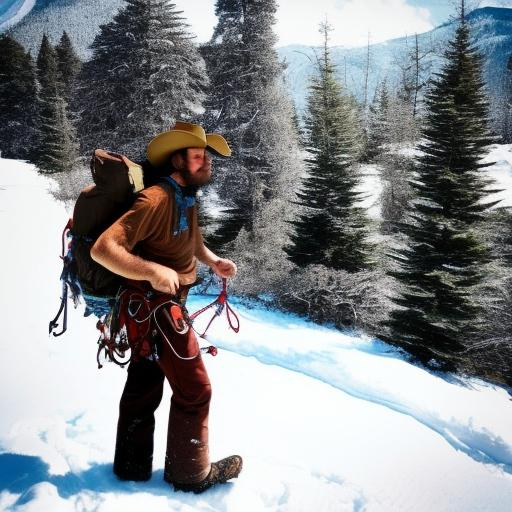}
    \end{minipage}%
    \begin{minipage}[t]{0.13\textwidth}
        \centering
        \includegraphics[width=\textwidth]{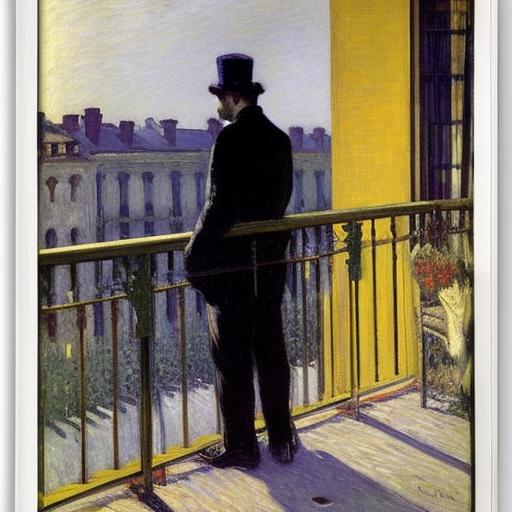}
    \end{minipage}%
    \begin{minipage}[t]{0.13\textwidth}
        \centering
        \includegraphics[width=\textwidth]{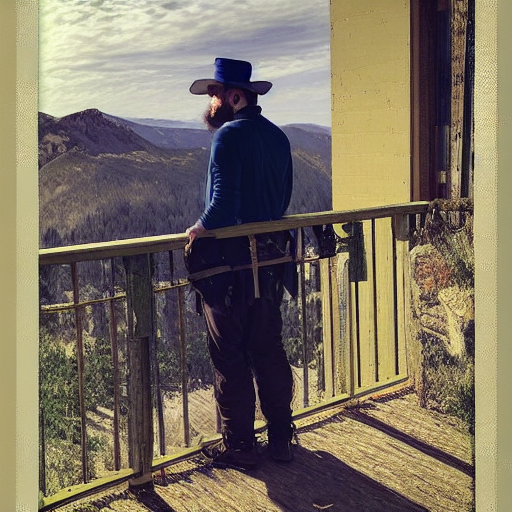}
    \end{minipage}%
    \begin{minipage}[t]{0.13\textwidth}
        \centering
        \includegraphics[width=\textwidth]{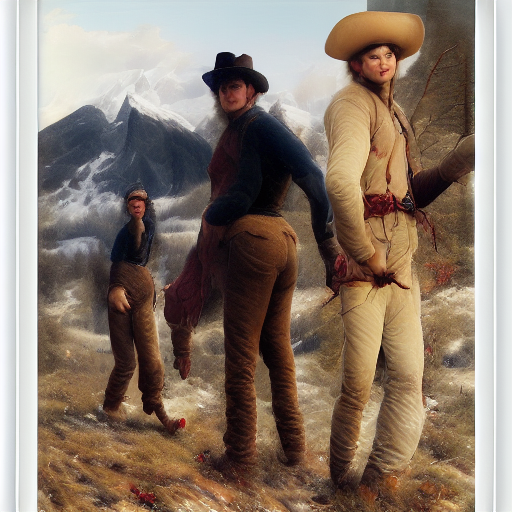}
    \end{minipage}%
    \begin{minipage}[t]{0.13\textwidth}
        \centering
        \includegraphics[width=\textwidth]{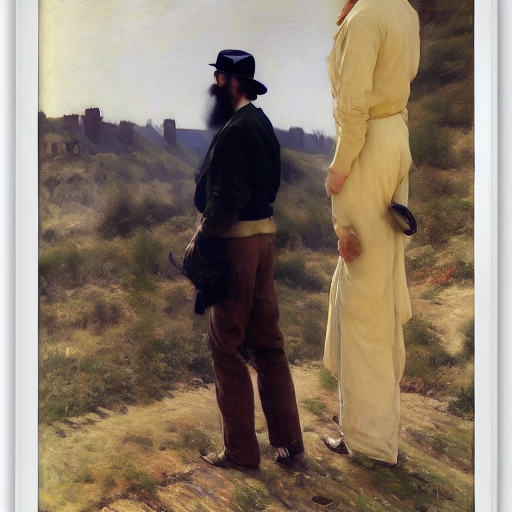}
    \end{minipage}%
    \begin{minipage}[t]{0.13\textwidth}
        \centering
        \includegraphics[width=\textwidth]{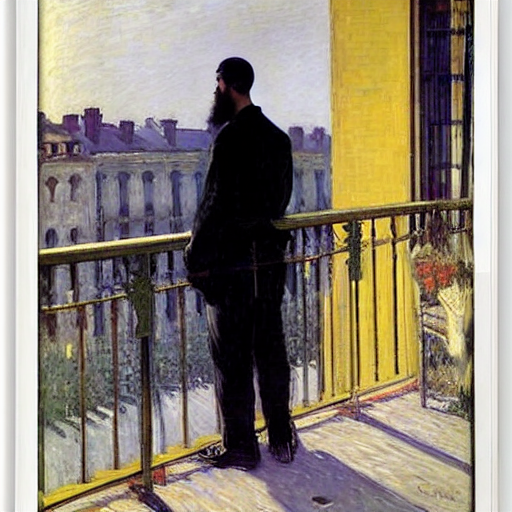}
    \end{minipage}%
    \par

    \begin{minipage}[t]{0.13\textwidth}
        \centering
        \includegraphics[width=\textwidth]{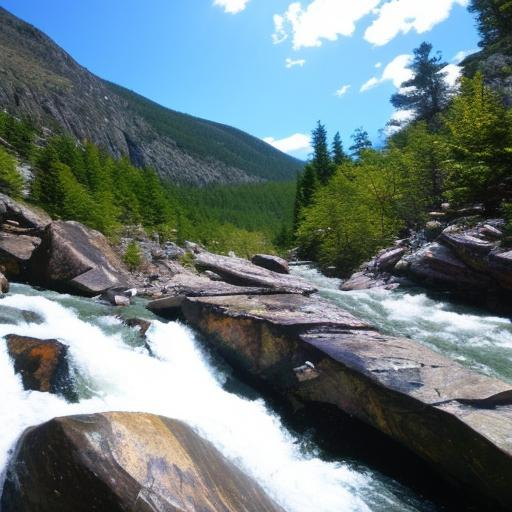}
    \end{minipage}%
    \begin{minipage}[t]{0.13\textwidth}
        \centering
        \includegraphics[width=\textwidth]{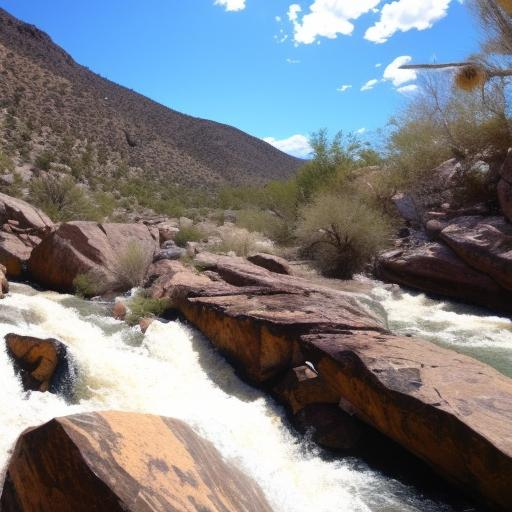}
    \end{minipage}%
    \begin{minipage}[t]{0.13\textwidth}
        \centering
        \includegraphics[width=\textwidth]{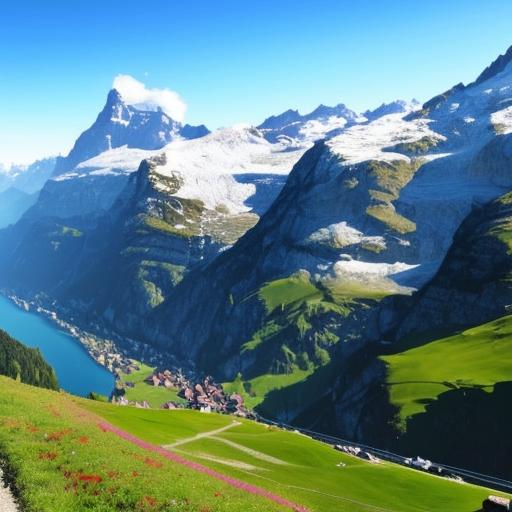}
    \end{minipage}%
    \begin{minipage}[t]{0.13\textwidth}
        \centering
        \includegraphics[width=\textwidth]{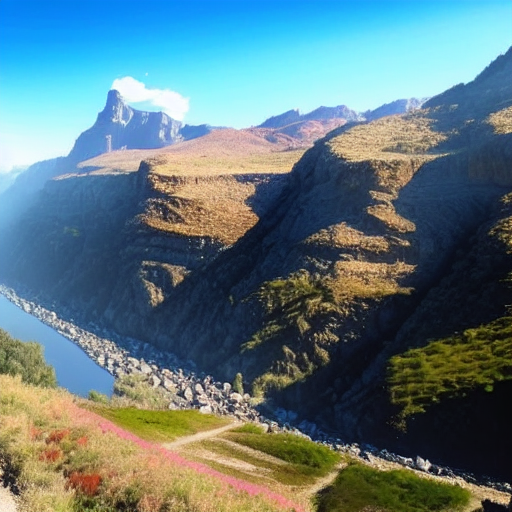}
    \end{minipage}%
    \begin{minipage}[t]{0.13\textwidth}
        \centering
        \includegraphics[width=\textwidth]{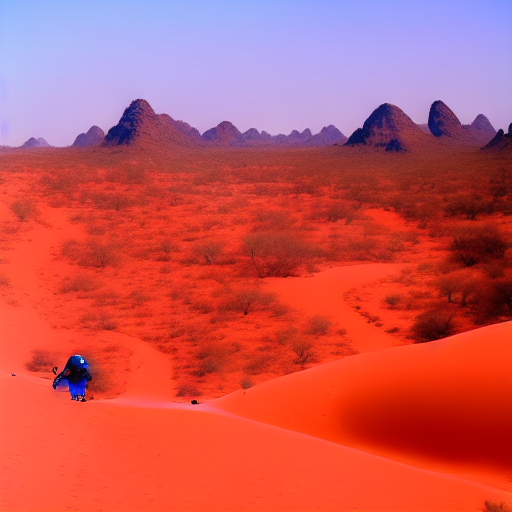}
    \end{minipage}%
    \begin{minipage}[t]{0.13\textwidth}
        \centering
        \includegraphics[width=\textwidth]{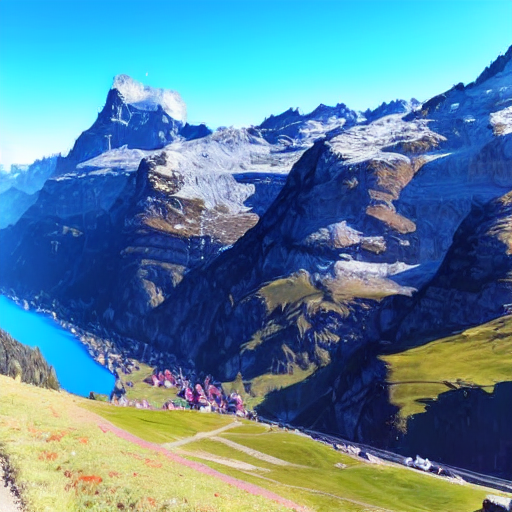}
    \end{minipage}%
    \begin{minipage}[t]{0.13\textwidth}
        \centering
        \includegraphics[width=\textwidth]{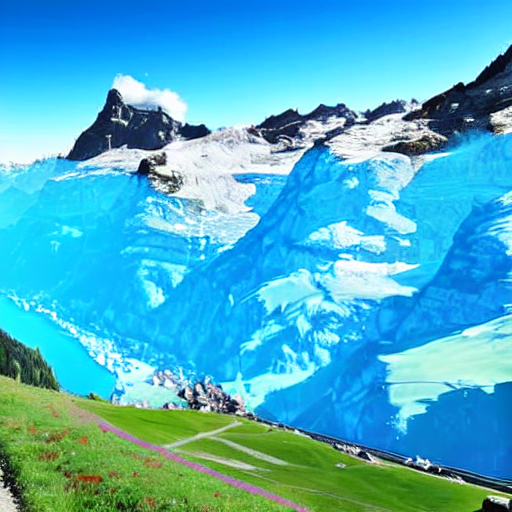}
    \end{minipage}%
    \par

    \begin{minipage}[t]{0.13\textwidth}
        \centering
        \includegraphics[width=\textwidth]{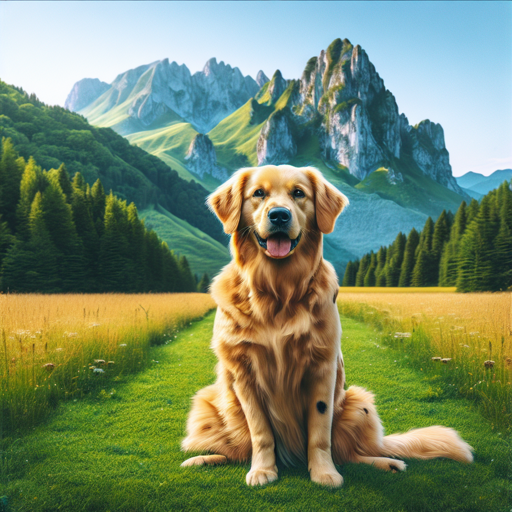}
    \end{minipage}%
    \begin{minipage}[t]{0.13\textwidth}
        \centering
        \includegraphics[width=\textwidth]{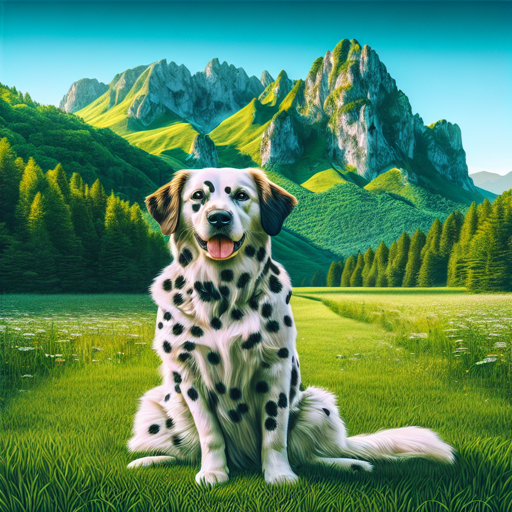}
    \end{minipage}%
    \begin{minipage}[t]{0.13\textwidth}
        \centering
        \includegraphics[width=\textwidth]{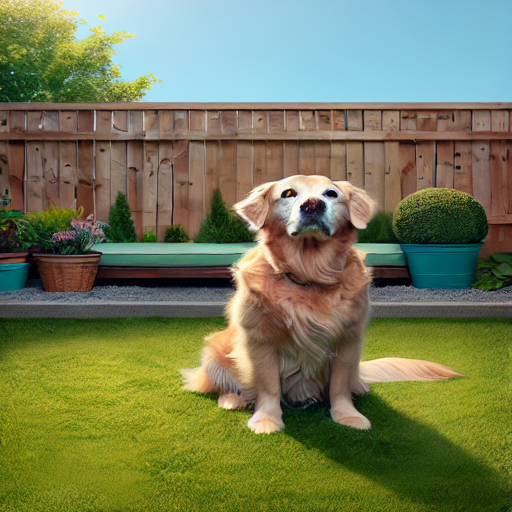}
    \end{minipage}%
    \begin{minipage}[t]{0.13\textwidth}
        \centering
        \includegraphics[width=\textwidth]{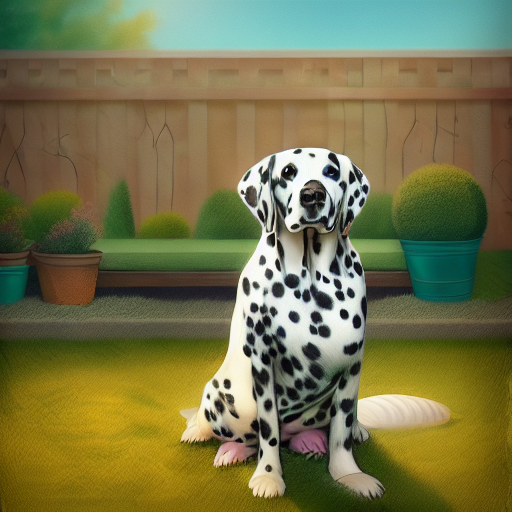}
    \end{minipage}%
    \begin{minipage}[t]{0.13\textwidth}
        \centering
        \includegraphics[width=\textwidth]{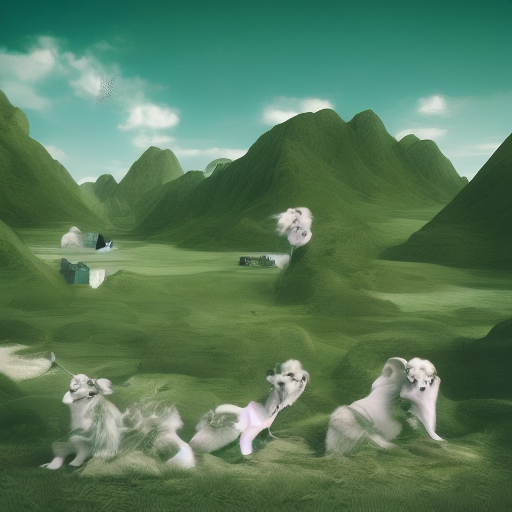}
    \end{minipage}%
    \begin{minipage}[t]{0.13\textwidth}
        \centering
        \includegraphics[width=\textwidth]{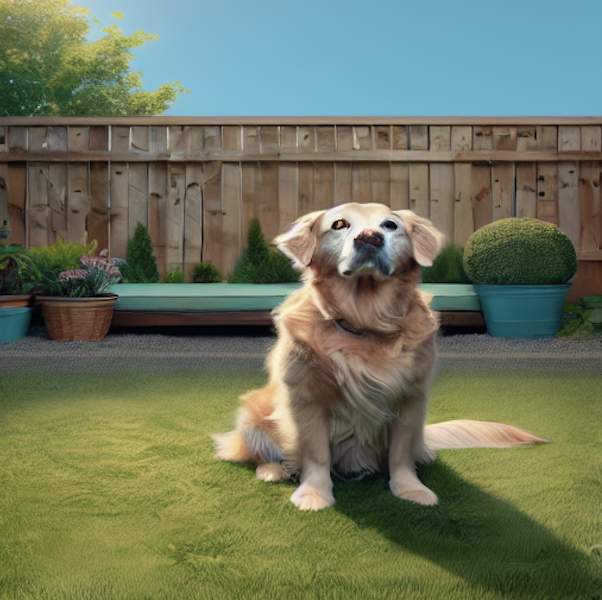}
    \end{minipage}%
    \begin{minipage}[t]{0.13\textwidth}
        \centering
        \includegraphics[width=\textwidth]{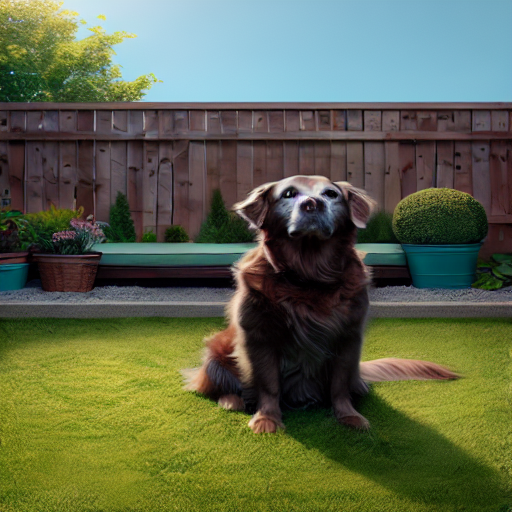}
    \end{minipage}%
    \par

    \begin{minipage}[t]{0.13\textwidth}
        \centering
        \includegraphics[width=\textwidth]{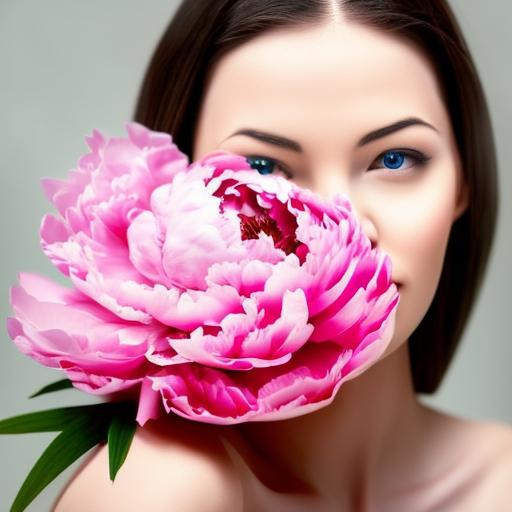}
    \end{minipage}%
    \begin{minipage}[t]{0.13\textwidth}
        \centering
        \includegraphics[width=\textwidth]{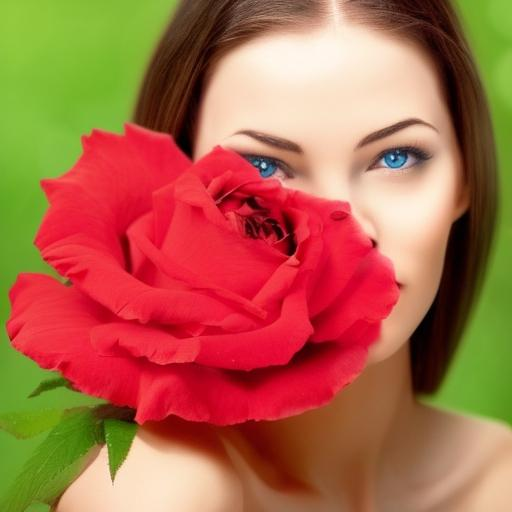}
    \end{minipage}%
    \begin{minipage}[t]{0.13\textwidth}
        \centering
        \includegraphics[width=\textwidth]{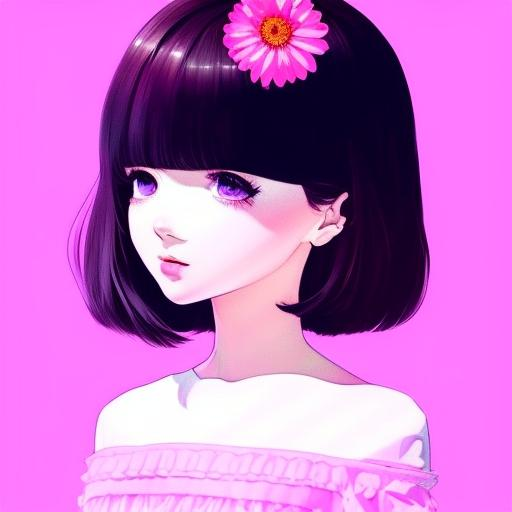}
    \end{minipage}%
    \begin{minipage}[t]{0.13\textwidth}
        \centering
        \includegraphics[width=\textwidth]{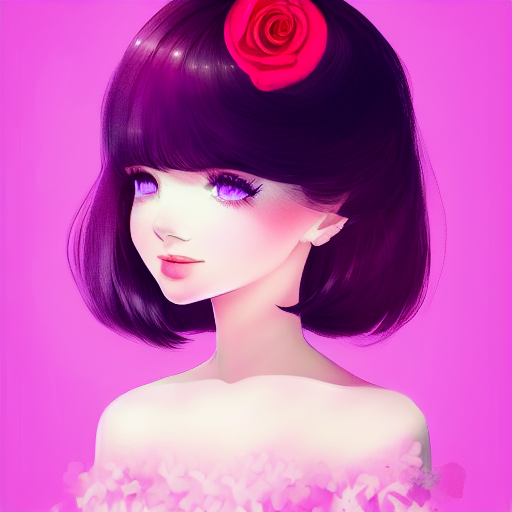}
    \end{minipage}%
    \begin{minipage}[t]{0.13\textwidth}
        \centering
        \includegraphics[width=\textwidth]{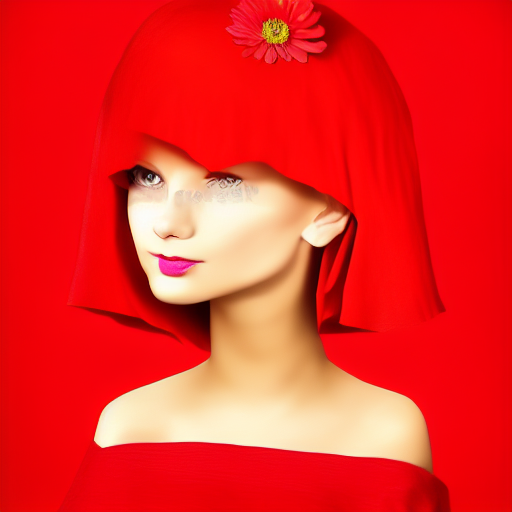}
    \end{minipage}%
    \begin{minipage}[t]{0.13\textwidth}
        \centering
        \includegraphics[width=\textwidth]{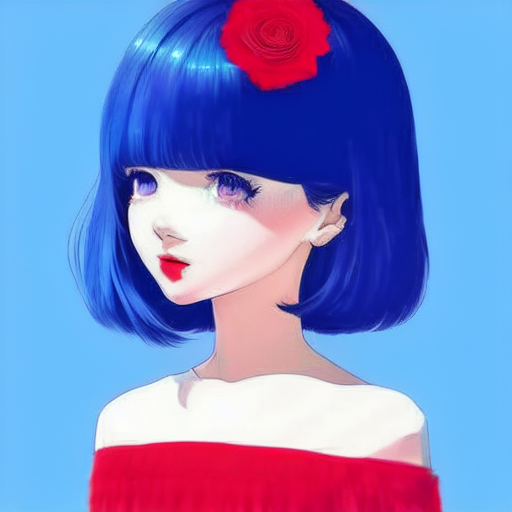}
    \end{minipage}%
    \begin{minipage}[t]{0.13\textwidth}
        \centering
        \includegraphics[width=\textwidth]{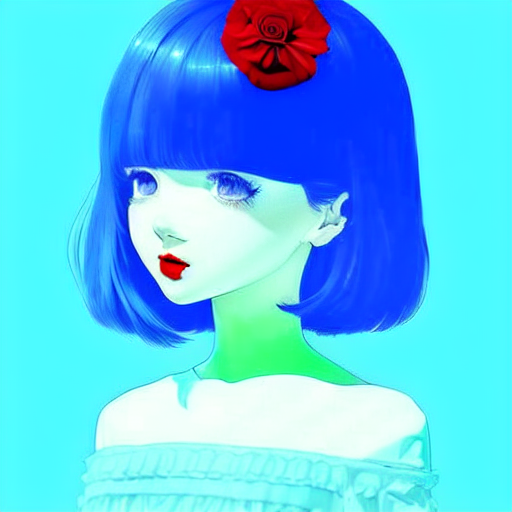}
    \end{minipage}%
    \par

    \begin{minipage}[t]{0.13\textwidth}
        \centering
        \includegraphics[width=\textwidth]{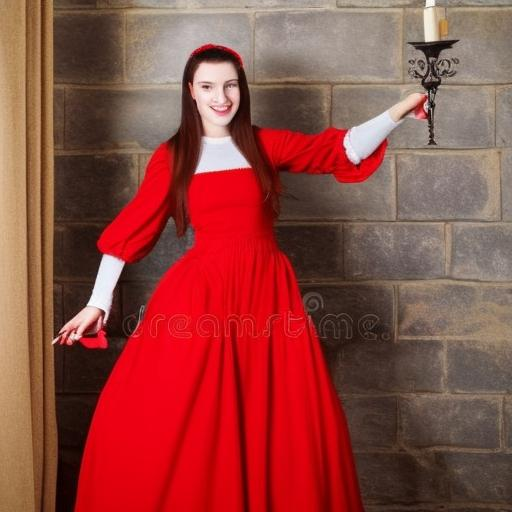}
    \end{minipage}%
    \begin{minipage}[t]{0.13\textwidth}
        \centering
        \includegraphics[width=\textwidth]{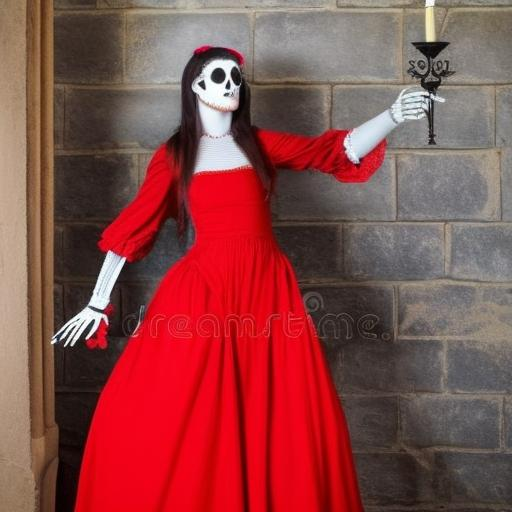}
    \end{minipage}%
    \begin{minipage}[t]{0.13\textwidth}
        \centering
        \includegraphics[width=\textwidth]{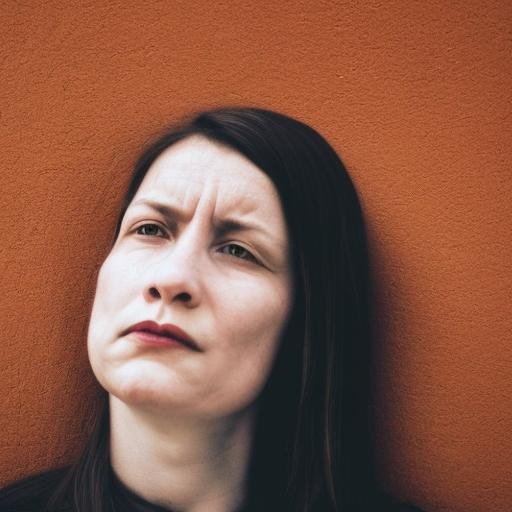}
    \end{minipage}%
    \begin{minipage}[t]{0.13\textwidth}
        \centering
        \includegraphics[width=\textwidth]{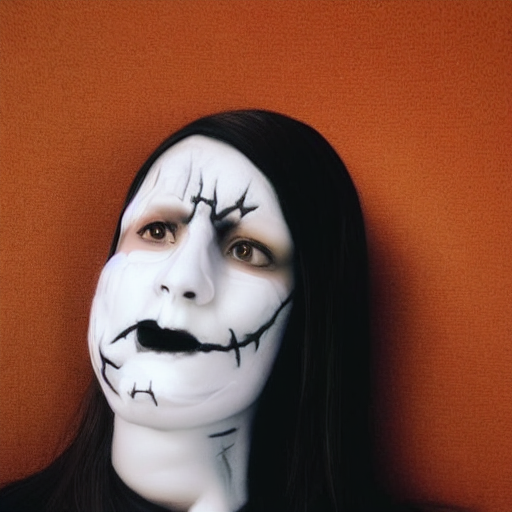}
    \end{minipage}%
    \begin{minipage}[t]{0.13\textwidth}
        \centering
        \includegraphics[width=\textwidth]{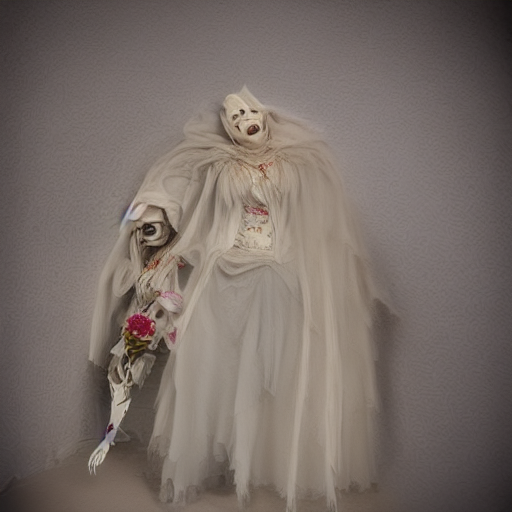}
    \end{minipage}%
    \begin{minipage}[t]{0.13\textwidth}
        \centering
        \includegraphics[width=\textwidth]{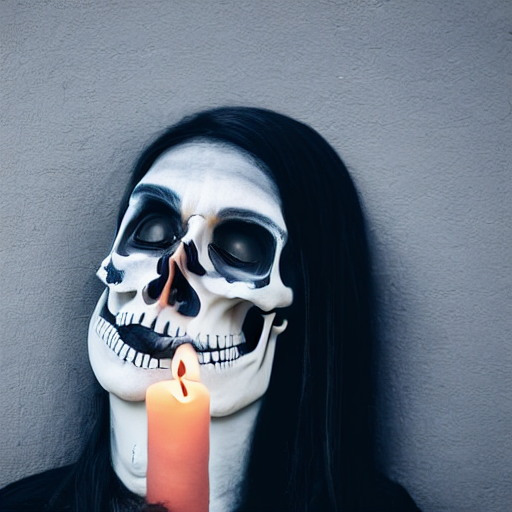}
    \end{minipage}%
    \begin{minipage}[t]{0.13\textwidth}
        \centering
        \includegraphics[width=\textwidth]{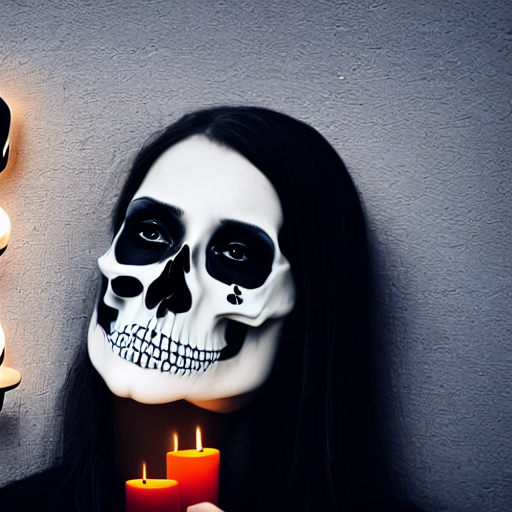}
    \end{minipage}%
    \par

    \caption{Qualitative comparison of our framework \method with strong baselines (VISII, Instruct-pix2pix) for exemplar-based image editing. \method consistently produces images with higher edit accuracy and better consistency in non-edited regions compared to the baselines. Zoom in for better view.\label{fig:qual}}
\end{figure}

We support our quantitative findings with qualitative examples illustrated in Fig.~\ref{fig:qual}. In \textit{rows 1 and 6}, the face of the person is correctly changed to a skeleton without compromising the structure or fidelity of the rest of the image. In some cases, baselines VISII \& IP2P do not capture the edit correctly and their outputs suffer from unwanted artifacts (for e.g., addition of candle in \textit{row 6}). On the contrary, \method performs favorably on the difficult tasks of local edits (Golden Retriever to Dalmatian in \textit{row 4}, person to cowboy in \textit{row 2}) as well as global edits (change of scene to desert in \textit{row 3}).

\vspace{-0.45cm}
\section{Conclusion}
\label{sec:conclusion}
\vspace{-0.35cm}

In this paper, we present an optimization-free approach to edit an image following the edit in a given exemplar pair. To achieve this, we first capture the edit in both text and image spaces, and condition the denoising of a pretrained diffusion model on the captured edit, and features from original image to obtain the edited output. Our results show this is a cost-effective method and also transfers superior edits compared to baselines. Extension of this work to other text-to-image diffusion models beyond SD is part of our future work. We hope our findings motivate further research into exemplar-based image editing.


%
%
\newpage
\bibliographystyle{splncs04}
\bibliography{main}
\end{document}